\documentclass[11pt]{article}
\usepackage{amsmath,amssymb} 
\usepackage{graphicx}
\usepackage{bm}
\usepackage{mathrsfs}
\usepackage{subfigure}
\usepackage{amsfonts}
\usepackage{color}
\usepackage{yhmath}
\usepackage{amscd}
\usepackage{latexsym}
\usepackage{arydshln}
\usepackage{multirow}
\usepackage{slashbox}
\begin{document}

\title{Unconstrained Two-parallel-plane Model for Focused Plenoptic Cameras Calibration}

\author{Chunping Zhang, Zhe Ji, Qing Wang\\
School of Computer Science\\
Northwestern Polytechnical University\\
qwang@nwpu.edu.cn}

\maketitle

\setlength\arraycolsep{4.0pt}

\begin{abstract}
The plenoptic camera can capture both angular and spatial information of the rays, enabling 3D reconstruction by single exposure. The geometry of the recovered scene structure is affected by the calibration of the plenoptic camera significantly. In this paper, we propose a novel unconstrained two-parallel-plane (TPP) model with 7 parameters to describe a 4D light field. By reconstructing scene points from ray-ray association, a 3D projective transformation is deduced to establish the relationship between the scene structure and the TPP parameters. Based on the transformation, we simplify the focused plenoptic camera as a TPP model and calibrate its intrinsic parameters. Our calibration method includes a close-form solution and a nonlinear optimization by minimizing re-projection error. Experiments on both simulated data and real scene data verify the performance of the calibration on the focused plenoptic camera.
\end{abstract}

\section{Introduction}

The micro-lens array (MLA) based plenoptic cameras, including the conventional plenoptic camera \cite{ng2006digital} and the focused plenoptic camera \cite{lumsdaine2009focused}, capture the radiance information of rays in both spatial and angular dimensions, e.g. the 4D light field data \cite{levoy1996light,gortler1996lumigraph}. The data from the plenoptic camera is equivalent to narrow baseline images of traditional cameras with projection centers on the lens aperture plane. The measurement of the same position in multiple directions allows or strengths applications on computer photography, such as digital refocusing \cite{ng2005fourier}, depth estimation \cite{Jeon2015Accurate}, saliency detection \cite{li2014saliency} and so on. However, the angular and spatial resolution of a light field data is limited by the physical parameters of the plenoptic camera. Recent work proposed the methods on light field registration and stitching to expand the field of view \cite{johannsen2015linear,birklbauer2014panorama,guo2015enhancing}. To support these applications, calibrating the plenoptic camera and decoding accurate 4D light field in metric distance from the 2D image sensor are crucial.

To calibrate the plenoptic camera, it is essential to build a model to relate the measurement on the 2D raw image and the rays in the 3D space. Prior work dealt with the intrinsic calibration on the plenoptic cameras in different optical designs \cite{dansereau2013decoding,bok2014geometric,heinze2015automated,vaish2004using}. However, the parameters of the proposed models are redundant or incomplete, and the models are still improvable on the description of plenoptic cameras. Some of the calibration methods have issues on the initialization estimation or the optimization procedure.

In this paper, we propose a novel unconstrained TPP model with 7 parameters to describe the light field structure inside and outside the plenoptic camera concisely. The 7 parameters are sufficient to constrain the rays in a 4D light field. Based on the 7-parameter TPP model, the pixels on the raw image can be related to the rays by a virtual MLA directly. We deduce the projective transformation \cite{hartley2003multiple} on the reconstructed 3D points with different TPP parameters, which is the theoretical foundation of the closed-form solution of our calibration method. Then we employ a nonlinear optimization to refine the parameters via minimizing the re-projection error on the raw images. We conduct experiments on both simulated data and a physical plenoptic camera.

In summary, our main contributions are listed as follows:

(1) We simplify the plenoptic camera system as a 7-parameter unconstrained TPP coordinate and deduce a projective transformation matrix to relate the measurement on the image sensor to the scene points in 3D space.

(2) We solve the parameters of the TPP using a robust and efficient method, which consists of a linear initialization and an optimization via re-projection error.

The remainder of this paper is organized as follows: Section~\ref{sec:RelatedWork} summarizes the related work on the plenoptic camera models and calibration methods. Section~\ref{sec:TPPModel} describes the 7-parameter unconstrained TPP model and its relationship with the physical plenoptic camera, and derives the projective transformation involved the TPP's parameters. Section~\ref{sec:Solve} provides the details of our proposed calibration method. Section~\ref{sec:ExpResults} shows the calibration results on simulated data and real scene data.

\section{Related Work}
\label{sec:RelatedWork}

To acquire light field, there are various imaging systems developed from the traditional camera. Wilburn et al. \cite{wilburn2005high} presented a camera array to obtain light field with high spatial and angular resolution. Prior work dealt with the calibration of the camera arrays \cite{vaish2004using}. Unfortunately, applications on camera arrays are limited by its high cost and complex control. In contrast, a MLA enables a single camera to record 4D light field more conveniently and efficiently, though the baseline and spatial resolution is smaller than the camera array. Recent work devoted to calibrate the intrinsic parameters of the plenoptic cameras in two designs \cite{ng2006digital,lumsdaine2009focused}, which are quite different according to the image structure of the micro lenses. Moreover, in traditional multi-view geometry, multiple cameras in different poses are defined as a set of unconstrained rays, which is known as as Generalized Camera Model (GCM) \cite{pless2003using}. The ambiguity of the reconstructed scene was discussed in traditional topics. For a plenoptic camera, the different views of the same scene point are obtained, and the calibration of a plenoptic camera can use the theory on traditional multi-view for reference.

Some work explored the calibration on the focused plenoptic camera, where the multiple projections of the same scene point are convenient to be recognized. Johannsen et al. \cite{johannsen2013calibration} proposed the method on the intrinsic parameters calibration of a focused plenoptic camera. By reconstructing 3D points from the parallax in adjacent micro-lens images, the parameters including the depth distortion were estimated by nonlinear optimization directly without a linear initialization. However, the geometry center of the micro image was on its micro-lens's optical axis in their method. This assumption caused inaccuracy on the reconstructed points and was compensated by the depth distortion coefficients. Hahne et al. \cite{hahne2015refocusing} discussed the influence of the deviation of the micro image's centers and the optical center of its micro lens. Heinze et al. \cite{heinze2015automated} applied a similar method with \cite{johannsen2015linear} and proposed a linear initialization for the intrinsic parameters.

Some work explored the calibration on the conventional plenoptic camera, where the sub-aperture images are easy to be synthesized. Dansereau et al. \cite{dansereau2013decoding} presented a model to decode the pixels into rays for a conventional plenoptic camera, where the 12-free-parameter transformation matrix was connected with the reference plane outside the camera. However, the calibration method was initialized using traditional camera calibration techniques and there were redundant parameters in the decoding matrix. Bok et al. \cite{bok2014geometric} formulated a geometric projection model consisting of a main lens and a MLA to estimate the intrinsic and extrinsic parameters by utilizing raw images directly, including an analytical solution and a nonlinear optimization. Moreover, Thomason et al. \cite{thomason2014calibration} concentrated on the misalignment of the MLA and estimated its position and orientation.

Different from the previous models, we represent the image sensor, the MLA and the main lens as a simple TPP model with 7 parameters. The 7 parameters are connected with the physical parameters of the plenoptic camera and sufficient to relate the pixels on the raw image to the rays without redundancy. To reveal the relationship between the light field data and the scene structure, we explore the ray-ray association on the TPP coordinate. Then the 3D projective transformation of the reconstructed structure with different TPP parameters is deduced. Based on the projective transformation, we solve a linear initialization for the intrinsic and extrinsic parameters. In our method, the prior scene points are support by a planar calibration board in different poses. The solved initialization is refined by minimizing the re-projection error using Levenberg-Marquardt algorithm. Theoretical derivation and experimental results demonstrate the validity of our calibration method.

\section{Unconstrained TPP model}
\label{sec:TPPModel}

\begin{figure}
\centering
\includegraphics[width=60mm]{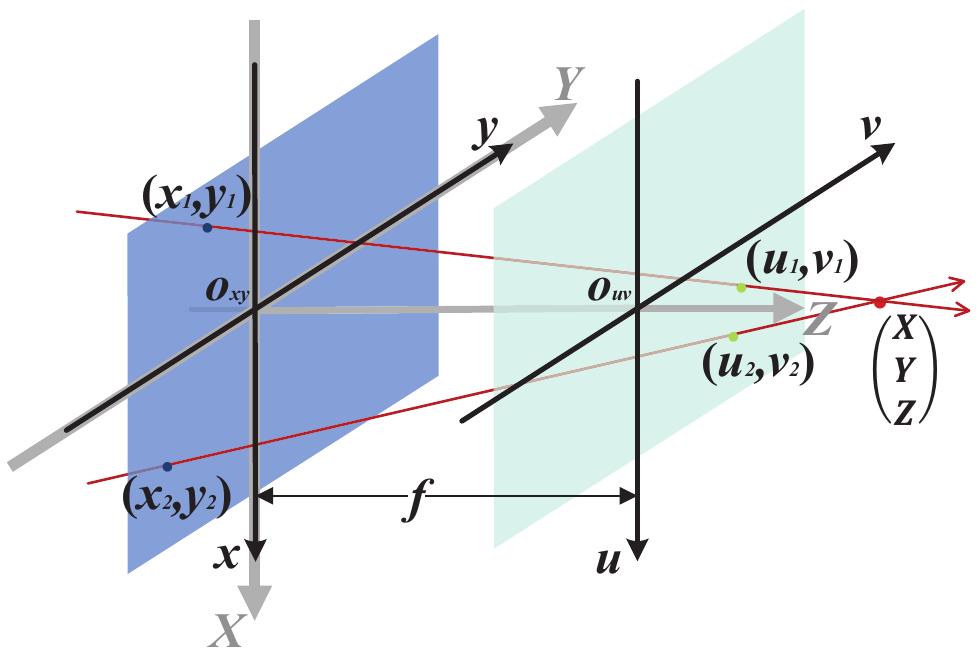}
\vspace{-3mm}
\caption{
An illustration of TPP coordinate system. The origins of two coordinates $x$-$y$ and $u$-$v$ lie on the axis $Z$. Axis $x$ and axis $y$ are parallel to axis $u$ and axis $v$ respectively. The distance between $x$-$y$ plane and $u$-$v$ plane is $f$.}
\label{fig:TPP_coordinate}
\vspace{-3mm}
\end{figure}

The distance of the traditional TPP model is normalized as 1 unit to describe a set of rays \cite{levoy1996light,gortler1996lumigraph}. To describe the decoded rays of a plenoptic camera, a TPP model with free parameters are needed, e.g. the unconstrained TPP model. As shown in Fig.\ref{fig:TPP_coordinate}, we define a TPP coordinate, where $\vec{r}\!=\! \left( x, y, u, v, f \right)\!\! ^\mathrm{T}$ defines a ray passing $\left(x, y, 0 \right)\!\! ^\mathrm{T}$ and $\left(u, v, f \right)\!\! ^\mathrm{T}$. In this section, we discuss the 3D projective transformation of reconstructed points in light field based on the TPP model. Then we establish the relationship between the TPP parameters and physical parameters of a focused plenoptic camera.

\subsection{Projective Transformation on TPP}
\label{subsec:tpp}

Let $\vec{r}$ pass the point $\left(X, Y, Z\right)^\mathrm{T}$, we have:

\vspace{-2mm}
\begin{equation}
   \underbrace{ \left[  \begin{array}{cccc}
  f & 0 & x\!-\!u & -\!fx \\
  0 & f & y\!-\!v & -\!fy \end{array} \right] }_{\bm{M}}
  \left[ \begin{array} {c}
  	X \\ Y \\ Z \\ 1 \end{array} \right] = \bm{0},
  \label{eq_Mx=0}
\end{equation}

\noindent where $\left(X, Y, Z\right)^\mathrm{T}$ can be solved iff there are at least two rays and any two of the rays $\vec{r}_i$ and $\vec{r}_j$ satisfy $\frac{u_i-u_j}{x_i-x_j}=\frac{v_i-v_j}{y_i-y_j}$.

\begin{figure}
\centering
\includegraphics[width=90mm]{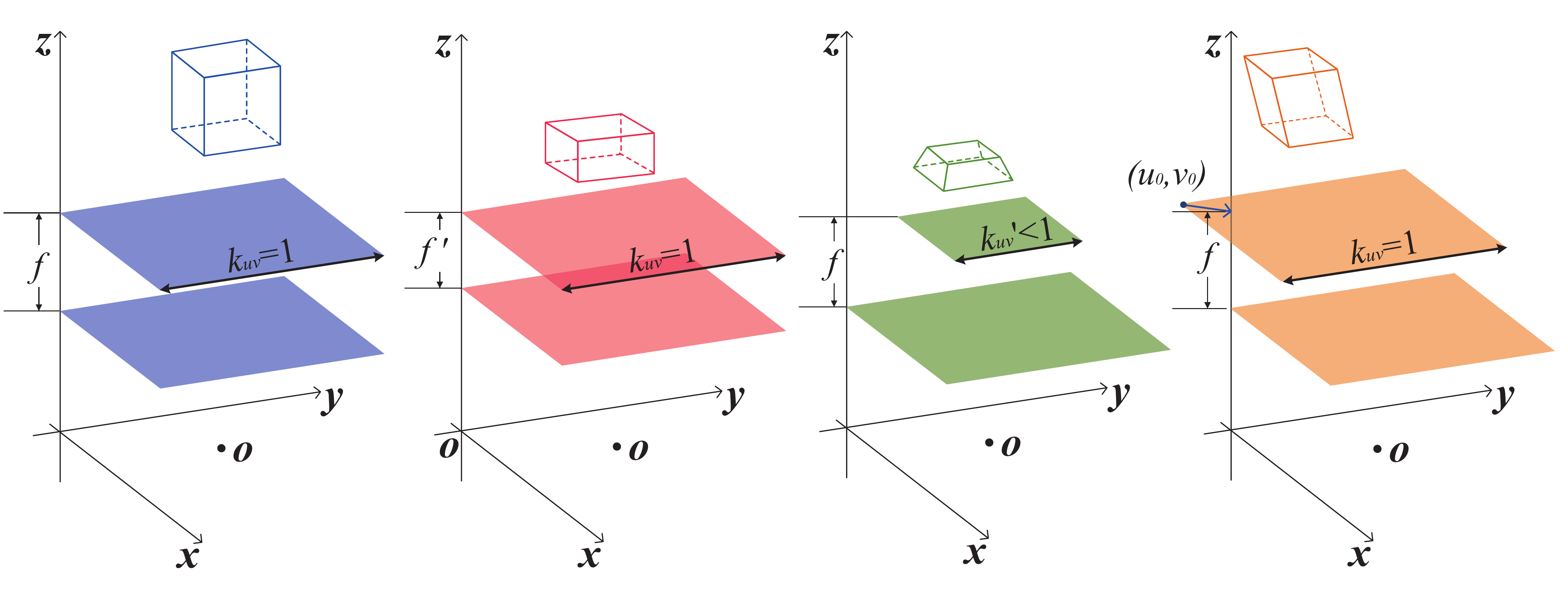}
\vspace{-4mm}
\caption{
TPP light field recording a Lambertian cube. The left one is the original cube and the others are distorted cubes with the change of $f$, $k_{u\!v}\left(k_{uv}\!\!=\!\! k_u\!\!=\!\! k_v\right)$, $\left(u_0, v_0 \right)^\mathrm{T}$ respectively.}
\label{fig:projective_distortion}
\vspace{-3mm}
\end{figure}

Transforming $\vec{r}$ into $\vec{r}^\prime \!\!=\!\! \left( k_{\!x}x, k_{\!y}y, k_{\!u}u\!+\!u_0, k_{\!v}v\!+\!v_0, f\prime \right)\!\! ^\mathrm{T}$, the intersection point $\left(X, Y, Z \right)\!\!^\mathrm{T}$ is changed to be $\left(X^\prime, Y^\prime, Z^\prime \right)\!\! ^\mathrm{T}$, which satisfies:

\vspace{-2mm}
\begin{equation}
  \setlength\arraycolsep{4.0pt}
  \underbrace{ \left[  \begin{array}{cccc}
  fk_uk_x & 0 & k_x u_0 & 0 \\
  0 & fk_vk_x & k_x v_0 & 0 \\
  0 & 0 & f^\prime k_x & 0 \\
  0 & 0 & k_x\!-\!k_u & fk_u \end{array} \right] }_{=:P\left( \bm{\mathscr{X}}, f \right)}
  \left[\!\! \begin{array} {c}
  	X \\ Y \\ Z \\ 1 \end{array} \!\!\right] = s
  	\left[\!\! \begin{array} {c}
  	X^\prime \\ Y^\prime \\ Z^\prime \\ 1 \end{array} \!\!\right],
  \label{eq_PX=sX_}
\end{equation}
\vspace{-1mm}

\noindent where $\mathscr{X}\!\!\!=\!\! \left( k_x, k_y, k_u, k_v, u_0, v_0, f^\prime \right)$ is the transformation parameters of the TTP coordinate and $s$ is a scalar factor. To make the set of transformed rays $\vec{r}^\prime$ all intersect at the point $\left(X^\prime, Y^\prime, Z^\prime \right)^\mathrm{T}$, $\mathscr{X}$ must satisfy $k_u/k_x\!=\! k_v/k_y$. Equation \ref{eq_PX=sX_} indicates that $\mathscr{X}$ affects the geometric structure of the recovered scene, {\it i.e.} the intersections of rays. In addition, the non-zero elements in the last row of $\bm{P}$ is equivalent to the refraction of a lens in a traditional camera. Therefore, by transforming the coordinate of the TPP via $\mathscr{X}$, the light field inside a camera can be transformed into the real world scene, where the scale of $u$-$v$ handles the projective transformation and the scale of $x$-$y$ handles the zoom of the recovered scene. The transformations on the scene structure with single parameter in $\mathscr{X}$ are shown in Fig.\ref{fig:projective_distortion} separately.

In Section~\ref{sec:Solve}, we will discuss the calibration method using the projective transformation in Eq.\ref{fig:projective_distortion}.

\subsection{TPP Coordinate Inside and Outside the Camera}
\label{subsec:tpp2}

\begin{figure}
\centering
\includegraphics[width=80mm]{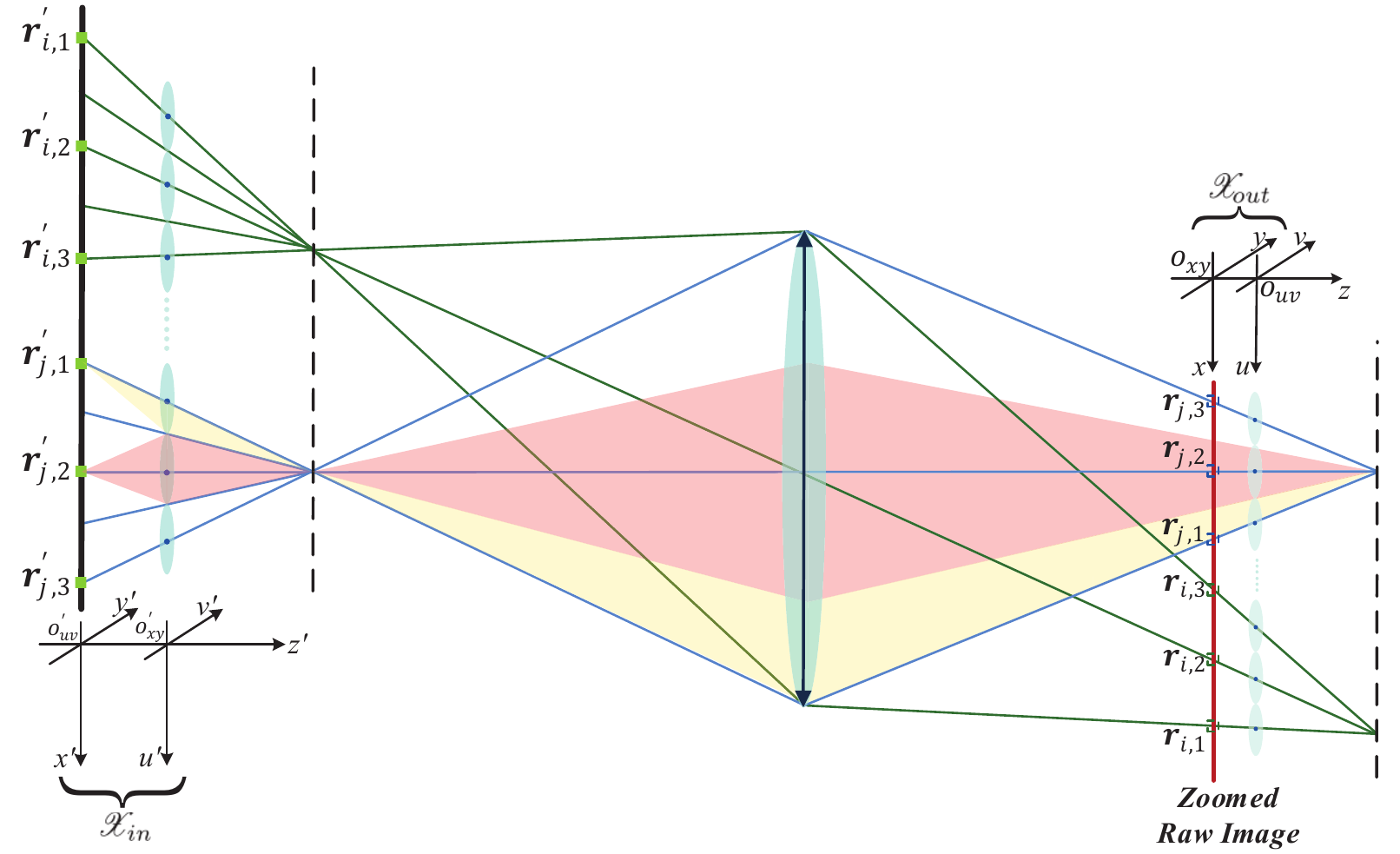}
\vspace{-3mm}
\caption{
A focused plenoptic camera with a MLA. There are two TPP coordinates, {\it i.e.} $x^\prime y^\prime u^\prime v^\prime$ inside the camera and $xyuv$ in the real world scene.}
\label{fig:TPP2System}
\end{figure}

We model the main lens as a thin lens and the micro-lens as a pinhole, thus every pixel on the image sensor can be regarded as a ray passing through the coordinate on the image sensor and the optical center of its corresponding micro-lens \cite{dansereau2013decoding,johannsen2013calibration,bok2014geometric}.
The TPP coordinate system consists of a image sensor and a MLA, {\it i.e.} the $x^\prime y^\prime u^\prime v^\prime$ coordinate shown in Fig.\ref{fig:TPP2System}. Moreover, there is another TPP coordinate $xyuv$ outside the camera, where the $x$-$y$ is related to the image sensor and the $u$-$v$ is related to the MLA. The $x$-$y$ and $u$-$v$ planes can be regarded as a zoomed raw image and a virtual MLA with larger diameter respectively.

Due to the refraction of the main lens, the rays and the scene structure in the two TPP coordinates are different. Obviously, there is a projective transformation on the reconstructed 3D points with different TPP parameters (Eq.\ref{eq_PX=sX_}). By transforming the rays $\vec{r}^\prime$ in $x^\prime y^\prime u^\prime v^\prime$ coordinate to $\vec{r}$ in $xyuv$ coordinate via $\mathscr{X}$, {\it i.e.} reparameterizing the coordinate of TTP, the projective distortion of the reconstructed points can be rectified. Therefore, the complex compositions of the plenoptic camera can be equivalently replaced by two parallel planes. In other words, there is a similar two-plane light field structure in the real world scene with the one inside the camera. Then we discuss the relationship between the physical structure inside a focused plenoptic camera and the virtual structure outside the camera.

In the plenoptic camera, a ray which passes the pixel $\left(x, y \right)$ on the image coordinate and the micro-lens with label $\left(i, j \right)$ can be represented as a virtual ray $\left(x, y, i, j, 1 \right)^{\mathrm{T}}$ where $i \! \in \! \mathbb{Z}$, $j \! \in \! \mathbb{Z}$. There is a geometric relationship between the two coordinates $xyuv$ and $x^\prime y^\prime u^\prime v^\prime$. Let $\mathscr{X}_{in}\!\!=\!\!\left( k_{x\!,in},k_{y\!,in},k_{u\!,in},k_{v\!,in}, u_{in}, v_{in}, f_{\!in} \right)\!\! ^\mathrm{T}$ and $\mathscr{X}_{\!out}\!\!=\!\!( k_{x\!,out},k_{y\!,out},k_{u\!,out},k_{v\!,out}, u_{out}$, $v_{out}, f_{\!out} )\!^\mathrm{T}$ be the parameters of TPP inside and outside the camera respectively, thus the virtual ray $\left(x, y, i, j, 1 \right)\!\! ^\mathrm{T}$ is related to two physical rays $\left(k_{x,in}x, k_{y,in}y, k_{u,in}i+\!u_{in}, k_{v,in}j+\!v_{in}, f_{in} \right)\! ^\mathrm{T}$ and $(k_{x,out}x,\!k_{y,out}y, k_{u,out}i+u_{out}$, $k_{v,out}j+v_{out}, f_{out} )^\mathrm{T}$respectively. The parameters $k_{x\!,in}$ (or $k_{y\!,in}$), $f_{in}$ can be regarded as the diameter of micro lens and the distance between the image sensor and the MLA. With the image sensor origin $\left( X_{os}, Y_{os}, Z_{os} \right)\!\!^{\mathrm{T}}$  and the reference micro-lens $\left( X_{oa}, Y_{oa}, Z_{oa} \right)\!\! ^{\mathrm{T}}$ whose label is $\left(0,0\right)$ (in the main lens coordinate), we can get the relationship between $\mathscr{X}_{in}$ and $\mathscr{X}_{out}$:

\vspace{-1mm}
\begin{equation}
\vspace{-3.0pt}
\frac{k_{x,in}}{k_{x,out}} = \frac{k_{y,in}}{k_{y,out}} = \frac{F}{Z_{os}-F},	
\vspace{-3.0pt}
\end{equation}

\begin{equation}
\vspace{-3.0pt}
\frac{k_{u,in}}{k_{u,out}} = \frac{k_{v,in}}{k_{v,out}} = \frac{F}{Z_{oa}-F},
\vspace{-3.0pt}
\end{equation}

\begin{equation}
\vspace{-3.0pt}
\left[\!\! \begin{array} {c}
u_{out}-u_{in} \\v_{out}-v_{in}
\end{array} \!\!\right] = \frac{Z_{oa}}{F-Z_{oa}}
\left[\!\! \begin{array} {c}
X_{oa} \\ Y_{oa}
\end{array} \!\!\right] - \frac{Z_{os}}{F-Z_{os}}
\left[\!\! \begin{array} {c}
X_{os} \\ Y_{os}
\end{array} \!\!\right]
\vspace{-3.0pt}
\end{equation}

\begin{equation}
\vspace{-1.0pt}
f_{in} = Z_{oa}\!\!-\!Z_{os},\quad
f_{out} = \frac{F}{F\!\!-\!Z_{oa}} Z_{oa}- \frac{F}{F\!\!-\!Z_{os}}Z_{os},\\
\vspace{-0.0pt}
\end{equation}

In addition, to simplify the discussion, we assume that the layout of the micro-lens array is square-like. For hexagon-like configuration, the label of micro-lens is different due to the layout.

\section{Calibration Method}
\label{sec:Solve}

To decode the virtual ray $\left(x, y, i, j, 1 \right)^{\!\mathrm{T}}$ into the ray in the real world scene, we need to calibrate the parameters of TPP coordinate system, {\it i.e.} the intrinsic parameters of the plenoptic camera. The theorem in Section~\ref{sec:TPPModel} indicates that given an arbitrary setting of TPP parameters, a set of 3D points can be recovered, and there is a projective transformation between the real scene points and the recovered points. Moreover, the projective transformation is determined by the TPP parameters.

This section provides the details of how to solve the parameters effectively, including a linear closed-form solution and a nonlinear optimization to minimize the re-projection error.

\subsection{Linear Initialization}
\label{subsec:linearInit}

Given an arbitrary setting $\mathscr{X}^\prime \!\!=\!\! \left( k_x^\prime, k_y^\prime, k_u^\prime, k_v^\prime, u_0^\prime, v_0^\prime, f^\prime \right)^\mathrm{T}$\!\!,
we decode the virtual ray $\left(x, y, i, j, 1 \right)^{\!\mathrm{T}}$\!\! into the ray passing
$\left( k_x^\prime x,k_y^\prime y, 0 \right)^{\mathrm{T}}$\!\! and $(k_u^\prime i + u_0^\prime,k_v^\prime j +v_0^\prime$,$f^\prime)^{\mathrm{T}}$. Then the distorted 3D point $\bm{X}_d$ can be reconstructed.
Obviously, there is a projective transformation between $\bm{X}_d$ and the real scene point $\bm{X}_c$ (Eq.\ref{eq_PX=sX_}). Therefore, using a transformation parameter setting $\mathscr{X}_d \!\!=\!\! \left(k_x, k_y, k_u, k_v, u_0, v_0, f^\prime \right)^\mathrm{T}$, $\bm{X}_c$ can be transformed to $\bm{X}_d$. Then we assume that the points in the world coordinate $\bm{X}_w$ is related to the TPP coordinate by a rigid motion, $\bm{X}_c \!=\! \bm{R}\bm{X}_w+\bm{t}$, with rotation $\bm{R} \in SO(3)$ and translation $\bm{t}\in \mathbb{R}^3 $. Let's denote the $i^\mathrm{th}$ column of $\bm{R}$ by $\bm{r}_i$. Here we assume that $k_x^\prime \!=\! k_y^\prime \!=\! k_{xy}^\prime$, $k_u^\prime \!=\! k_v^\prime \!=\! k_{uv}^\prime$, and the same as $\mathscr{X}$. The relationship between the $\bm{R}$, $\bm{t}$, $\bm{X}_w$, $\bm{X}_d$ and $\mathscr{X}_d$ is:

\vspace{-2mm}
\begin{equation}
s\bm{X}_d \!= \bm{P} \left( \mathscr{X}_d, f \right) \left[\!\!  \begin{array} {cccc}
\bm{r_1} & \bm{r_2} & \bm{r_3} & \bm{t} \\
0 & 0 & 0 & 1 \end{array} \!\!\right]
  \left[ \!\!\! \begin{array}{c}
X_w \\ Y_w \\ Z_w \\ 1 \end{array} \!\!\! \right],
\label{eq_sXd=PRtX}
\end{equation}

\noindent where $f$ is the distance of the calibrated two parallel planes. Obviously, there is a $4\times3$ homography matrix:

\vspace{-3mm}
\begin{equation}
\bm{H} \!=\! \bm{P} \left[ \begin{array} {ccc}
\bm{r}_1 & \bm{r}_2 & \bm{t} \\
0 & 0 & 1
\end{array}  \right].
\label{eq_H=Prt}
\end{equation}

We assume that the calibration board plane is $Z=0$ on the world coordinate, thus $Z_w=0$. Let's denote the $i^\mathrm{th}$ prior point by $\bm{X}_{\!w\!,i}\!=\!( X_{\!w\!,i}, Y_{\!w\!,i}, 1 )^{\!\mathrm{T}}$. Combining Eq.\ref{eq_Mx=0} and Eq.\ref{eq_sXd=PRtX}, we have:

\vspace{-1mm}
\begin{equation}
\bm{M}_i \, \bm{H} \left[ \!\! \begin{array}{c}
X_w \\ Y_w \\ 1	
\end{array}\!\! \right] = \bm{0},
\label{eq_MiHXw=0}
\end{equation}

\begin{equation}
\left[\! \begin{array}{c}
\bm{M}_1 \otimes \left( X_{w\!,1} \,\,\, Y_{w\!,1} \,\,\, 1 \right) \\
\vdots  \\
\bm{M}_n \otimes \left( X_{w\!,n} \,\,\, Y_{w\!,n} \,\,\, 1 \right)
\end{array} \!\right] \overrightarrow{\bm{H}}  = \bm{0},
\label{eq_MH=0}
\end{equation}

\noindent where $\bm{M}_i$ is a $2m_i\times4$ matrix which contains $\bm{X}_{w,i}$'s $m_i$ decoded rays from the raw image, $\overrightarrow{\bm{H}}$ is a $12\times1$ matrix stretched on row from $\bm{H}$, and $\otimes$ is a direct product operator. The homography $\bm{H}$ multiplied by an unknown factor can be estimated by Eq.\ref{eq_MH=0}. Let's denote the $i^\mathrm{th}$ column vector of $\bm{H}$ be $\bm{h}_i= \left(h_{1i}, h_{2i}, h_{3i}, h_{4i} \right)\!\! ^\mathrm{T}$. Utilizing the orthogonality and identity of $\bm{R}$, we have:

\vspace{-1mm}
\begin{equation}
\begin{aligned}
&\bm{h}_1^{\mathrm{T}} \bm{P}^{-\mathrm{T}} \bm{P}^{-1} \bm{h}_2 = 0, \\
&\bm{h}_1^{\mathrm{T}} \bm{P}^{-\mathrm{T}} \bm{P}^{-1} \bm{h}_1 = \bm{h}_2^{\mathrm{T}} \bm{P}^{-\mathrm{T}} \bm{P}^{-1} \bm{h}_2,
\end{aligned}
\label{eq_hPPh}
\end{equation}

\noindent where $\bm{P}\!=\! \bm{P}\left(\mathscr{X}_d, f \right)$. Let's denote $\bm{P}^{-\!\mathrm{T}} \!\bm{P}^{-\!1}$ by a symmetric matrix $\bm{Q}$, thus:

\vspace{-3mm}
\begin{equation}
\setlength{\arraycolsep}{4.0pt}
\bm{Q} = \left[\!\! \begin{array} {cccc}
\frac{1}{ f^{2} k_{\!x\!y}^{2}k_{\!u\!v}^{2} } & 0 & - \frac{u_0}{f^{\!\prime} \! f^{2} k_{\!x\!y}^{2}k_{\!u\!v}^{2}} & 0 \\
0 & \frac{1}{ f^{2} k_{\!x\!y}^{2}k_{\!u\!v}^{2} } & - \frac{v_0}{f^{\!\prime} \! f^{2} k_{\!x\!y}^{2}k_{\!u\!v}^{2}} & 0 \\
- \frac{u_0}{f^{\!\prime} \!   f^{2} k_{\!x\!y}^{2}k_{\!u\!v}^{2}} &\! -\frac{v_0}{\!f^{\!\prime} \! f^{2} k_{\!x\!y}^{2}k_{\!u\!v}^{2}} &  \frac{1}{f^{\!\prime 2}k_{\!x\!y}^2} \!+\! \frac{u_0^2+v_0^2}{f^{\!\prime2} \! f^{2} k_{\!x\!y}^{2}k_{\!u\!v}^{2}} \!+\! \frac{ \left( k_{\!u\!v}\!-\!k_{\!x\!y} \right) ^2 }{ f^{\!\prime 2}\! f^{2} k_{\!x\!y}^{2}k_{\!u\!v}^{2} }  & \frac{k_{\!u\!v}\!-k_{\!x\!y}}{ f^{\!\prime} \! f^{2} k_{\!x\!y} k_{\!u\!v}^{2} } \\
0 & 0 & \frac{k_{\!u\!v}-k_{\!x\!y}}{ f^{\!\prime} \! f^{2} k_{\!x\!y}k_{\!u\!v}^{2} } & \frac{1}{ f^{2}k_{\!u\!v}^{2} }
\end{array}	\!\!\right].
\label{eq_Q=PP}
\end{equation}

Note that there are only six distinct non-zero elements in $\bm{Q}$, denoted by $\bm{q}= \left(q_{11}, q_{13}, q_{23}, q_{33}, q_{34}, q_{44} \right)^\mathrm{T}$. To solve $\bm{Q}$, we have:

\vspace{-2mm}
\begin{equation}
\setlength{\arraycolsep}{8.0pt}
\left[ \begin{array}{cc}
h_{11}h_{12}+h_{21}h_{22} &  h_{11}^2\!\!-\!h_{12}^2\!+\!h_{21}^2\!\!-\!h_{22}^2\\
h_{11}h_{32}+h_{12}h_{31} &  2\left( h_{11}h_{31}\!\!-\!h_{12}h_{32} \right)\\
h_{21}h_{32}+h_{22}h_{31} &  2\left( h_{21}h_{31}\!\!-\!h_{22}h_{32} \right)\\
h_{31}h_{32}              &  h_{31}^2\!\!-\!h_{32}^2\\
h_{31}h_{42}+h_{32}h_{41} &  2\left( h_{31}h_{41}\!\!-\!h_{32}h_{42} \right)\\
h_{41}h_{42}              &  h_{41}^2\!\!-\!h_{42}^2\\
\end{array} \right]^{\!\!\mathrm{T}} \left[ \!\!
\begin{array}{c}
q_{11} \\ q_{13} \\ q_{23} \\ q_{33} \\ q_{34} \\ q_{44}
\end{array} \!\! \right] = \bm{0}.
\label{eq_hq=0}
\end{equation}

By stacking at least three such equations as Eq.\ref{eq_hq=0}, we will have in general a unique non-zeros solution for $\bm{q}$ denoted as $\hat{\bm{q}}= \left(\hat{q}_{11}, \hat{q}_{13}, \hat{q}_{23}, \hat{q}_{33}, \hat{q}_{34}, \hat{q}_{44}\right)\!\! ^\mathrm{T}$, which is defined up to an unknown scale factor $\lambda \left( \lambda \bm{q}=\hat{\bm{q}}\right)$. Once $\bm{Q}$ is estimated, we can solve all the parameter $\mathscr{X}_d$:

\begin{equation}
\begin{aligned}
& \lambda = f^{\prime 2} / \hat{q}_{11} \left[ \left(\hat{q}_{33}\hat{q}_{44}\!-\!\hat{q}_{34}^2 \right)-\hat{q}_{44}/\hat{q}_{11} \left( \hat{q}_{13}^2+\hat{q}_{23}^2 \right) \right], \\
& k_{xy} = \sqrt{ \hat{q}_{44} / \hat{q}_{11}} ,\\
& k_{uv} = \sqrt{ \hat{q}_{44} / \hat{q}_{11} } \left( 1+f^\prime \hat{q}_{34} / \hat{q}_{44} \right),\\
& u_0 = -f^\prime \hat{q}_{13} / \hat{q}_{11},\\
& v_0 = -f^\prime \hat{q}_{23} / \hat{q}_{11},\\
& f = \left( \sqrt{ \lambda / \hat{q}_{44} } \right) / k_{uv}.
\end{aligned}
\label{eq_linearSln}
\end{equation}

After solving $\mathscr{X}_d$, the virtual ray $\left(x,y,i,j,1\right)^\mathrm{T}$ is related to a ray passing the
$\left( k_{xy}^\prime / k_{xy} x, k_{xy}^\prime / k_{xy} y, 0 \right)^\mathrm{T}$ and 
$\left(k_{u\!v}^\prime / k_{u\!v} i \!+\! u_0^\prime \!-\! u_0, k_{u\!v}^\prime / k_{u\!v} j \!+\! v_0^\prime \!-\! v_0, f \right)^\mathrm{T}$
in the real world scene in metric distance, and the 3D points $\bm{X}_c$ are recovered. Let's denote the parameters of the TPP in the real scene world by $\mathscr{X}\!\!=\!\! \left( k_{x\!y}^\prime / k_{x\!y}, k_{x\!y}^\prime / k_{x\!y}, k_{u\!v}^\prime / k_{u\!v}, k_{u\!v}^\prime / k_{u\!v}, u_0^\prime\!\!-\! u_0,  v_0^\prime\!\!-\! v_0, f \right)\!^\mathrm{T}$. Then the extrinsic parameters $\bm{R}_i$, $\bm{t}_i$
for the $i^\mathrm{th}$ raw image is computed:

\vspace{-2mm}
\begin{equation}
\begin{aligned}
& \bm{r}_1 = \bm{P}^{-1} \bm{h}_1 / \left\| \bm{P}^{-1} \bm{h}_1 \right\|,\\
& \bm{r}_2 = \bm{P}^{-2} \bm{h}_2 / \left\| \bm{P}^{-2} \bm{h}_2 \right\|,\\
& \bm{r}_3 = \bm{r}_1 \times \bm{r}_2,\\
& \bm{t}\,\,\, = \bm{P}^{-1} \bm{h}_3 / \left\| \bm{P}^{-1} \bm{h}_1 \right\|.
\end{aligned}
\label{eq_r1r2r3}
\end{equation}

\subsection{Nonlinear Optimization}
\label{subsec:nonlinearOpt}

The optical property of the lenses and the physical machining error of MLA lead to the distortion of the rays. Distortion is in primarily radially symmetric due to the symmetric design of the plenoptic camera. Moreover, the $x$-$y$ plane and  $u$-$v$ plane are related to the image sensor and the MLA respectively. Therefore, we employ the radial distortion on the two coordinates of TPP \cite{Zhang2000A}:

\vspace{-2mm}
\begin{equation}
\begin{aligned}
& \hat{x}^d = (\hat{x}-x_c)(1+s_1r_{\!x\!y}^2+s_2r_{\!x\!y}^4)+x_c,& \\	
& \hat{y}^d = (\hat{y}-y_c)(1+s_1r_{\!x\!y}^2+s_2r_{\!x\!y}^4)+y_c,& \\
& \hat{u}^d = (\hat{u}-u_c)(1+t_1r_{\!u\!v}^2+t_2r_{\!u\!v}^4)+u_c,& \\
& \hat{v}^d = (\hat{v}-v_c)(1+t_1r_{\!u\!v}^2+t_2r_{\!u\!v}^4)+v_c,&
\end{aligned}
\label{eq_dirtortion}
\end{equation}

\noindent where $\left(x_c, y_c\right)\!\! ^\mathrm{T}$ and $\left(u_c, v_c\right)\!\! ^\mathrm{T}$ are the offsets as the origin of the distortion on two planes, 
$\left(\hat{x}^d, \hat{y}^d\right)\!\! ^\mathrm{T}$ and $\left(\hat{u}^d, \hat{v}^d\right)\!\! ^\mathrm{T}$ are the distorted points, $r_{\!xy}\!=\!\sqrt{\left(\hat{x}\!-\!x_c\! \right)^2 \!+\! \left(\hat{y}\!-\!y_c\! \right)^2}$ and $r_{\!uv}\!=\!\sqrt{\left(\hat{u}\!-\!u_c\! \right)^2 \!+\! \left(\hat{v}\!-\!v_c\! \right)^2}$. The parameters $s_i$ and $t_i$ are the distortion coefficients.

We minimize the following cost function with initialization solved in Section~\ref{subsec:linearInit} to refine the intrinsic and extrinsic parameters, including the distortion coefficients:

\vspace{-2mm}
\begin{equation}
\sum_{i=1}^{n}\!\sum_{j=1}^{p_i} { \left\| \bm{x}_{i,j} - \hat{\bm{x}}_{i,j}^d \left( \mathscr{X}, s_1, s_2, t_1, t_2, \bm{R}_i, \bm{t}_i, \bm{X}_{w,i} \right) \right\| },
\label{eq_re-projectionError}
\end{equation}

\noindent where $\bm{x}_{i,j}$ is the $j^\mathrm{th}$ projections of the prior scene point $\bm{X}_{w,i}$ on the image coordinate, and $p_i$ is the number of the projections of $\bm{X}_{w,i}$. In Eq.\ref{eq_re-projectionError}, $\bm{R}$ is parameterized by Rodrigues formula \cite{faugeras1993three}. Equation \ref{eq_re-projectionError} is the re-projection error in traditional computer vision. In addition, the Jacobian matrix of the cost function is simple and sparse. It can be solved with the LM Algorithm based on the trust region method \cite{madsen2004methods}. We use MATLAB's lsqnonlin function to complete the optimization.

\subsection{Summary}
\label{subsec:summary}

Our proposed calibration procedure is listed as follows:

\begin{enumerate}
\item Take at least 3 raw images with different poses of the calibration board by moving either the calibration board or the camera.
\item Detect the multiple projections corresponding to the scene points.
\item Calculate the $4\times3$ homography $\bm{H}_i$ for the $i^\mathrm{th}$ raw image via Eq.\ref{eq_MH=0}.
\item Estimate the intrinsic and extrinsic parameters via Eqs.\ref{eq_hq=0}, \ref{eq_linearSln} and \ref{eq_r1r2r3}.
\item Refine all the parameters via Eq.\ref{eq_re-projectionError} using LM Algorithm.
\end{enumerate}

\section{Experimental Results}
\label{sec:ExpResults}

In experiments, we apply our calibration method on the simulated data and the real world scene data. The prior scene points $\bm{X}_w$ are obtained by a planar calibration board with a circular grid pattern (Fig.\ref{fig:physicalCamera}). Due to the inevitable misalignment of the MLA and the image sensor, a preprocess on the raw image is needed (Section~\ref{subsec:rect}).

\subsection{Rectification}
\label{subsec:rect}

\begin{figure}
\centering
\includegraphics[width=50mm]{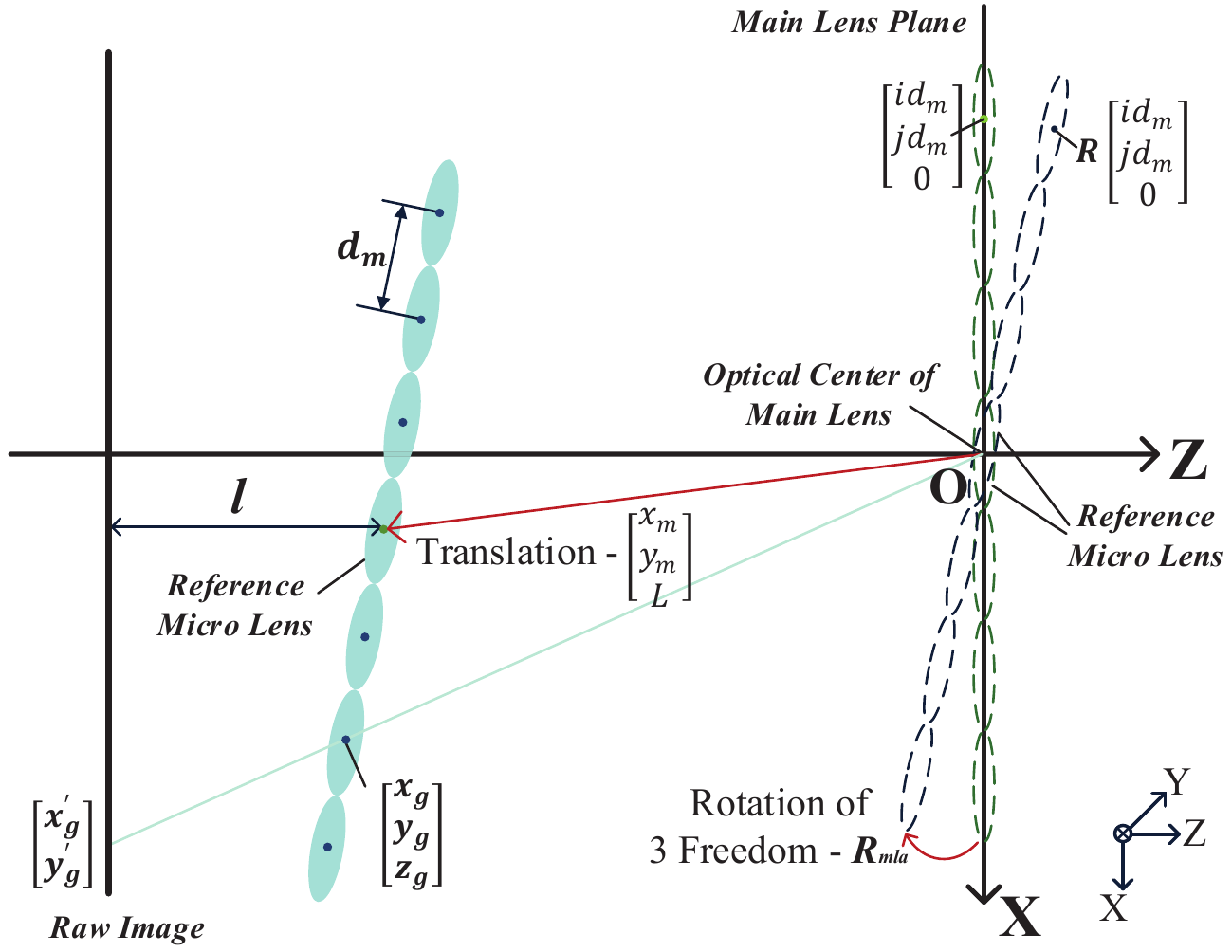}
\vspace{-3mm}
\caption{
The misalignment of the MLA in a plenoptic camera.}
\label{fig:misalignment_MLA}
\end{figure}

In a physical plenoptic camera, there is a slight rotation between the MLA and the image sensor \cite{thomason2014calibration}, as shown in Fig.\ref{fig:misalignment_MLA}. Let's denote the optical center and diameter of the micro-lens by $(x_g, y_g, z_g )^\mathrm{T}$ and $d_m$ respectively, we have:

\begin{equation}
\left[ \begin{array}{c;{2pt/2pt}c}
             & x_m \\	
\bm{R}_{m\!l\!a} & y_m \\	
             & L
\end{array} \right]	\left[\!\! \begin{array}{c}
id_{m} \\ jd_{m} \\ 0 \\ 1
\end{array} \!\!\right] = \left[\! \begin{array}{c}
x_g \\ y_g \\ z_g \\ 1	
\end{array} \!\right],
\label{eq_RtXg}
\end{equation}

\noindent where $\bm{R}_{mla} \!\!\in\!\! SO(3)$, and $\left( x_m,y_m,L \right)\!\! ^\mathrm{T}$ is the offset between the reference micro-lens and the main lens (Fig.\ref{fig:misalignment_MLA}). Therefore, the geometric center of the micro-lens image $(x_{g}^\prime, y_{g}^\prime)\! ^\mathrm{T}$ is:

\vspace{-1mm}
\begin{equation}
\left[\! \begin{array}{c}
x_g^{\,\prime}	\\ y_g^{\,\prime}
\end{array} \!\right] = \frac{L+l}{z_g} \left[\! \begin{array}{c}
x_g	\\ y_g
\end{array} \!\right].
\label{eq_xg_}
\end{equation}

\begin{figure}
\centering
\includegraphics[width=60mm]{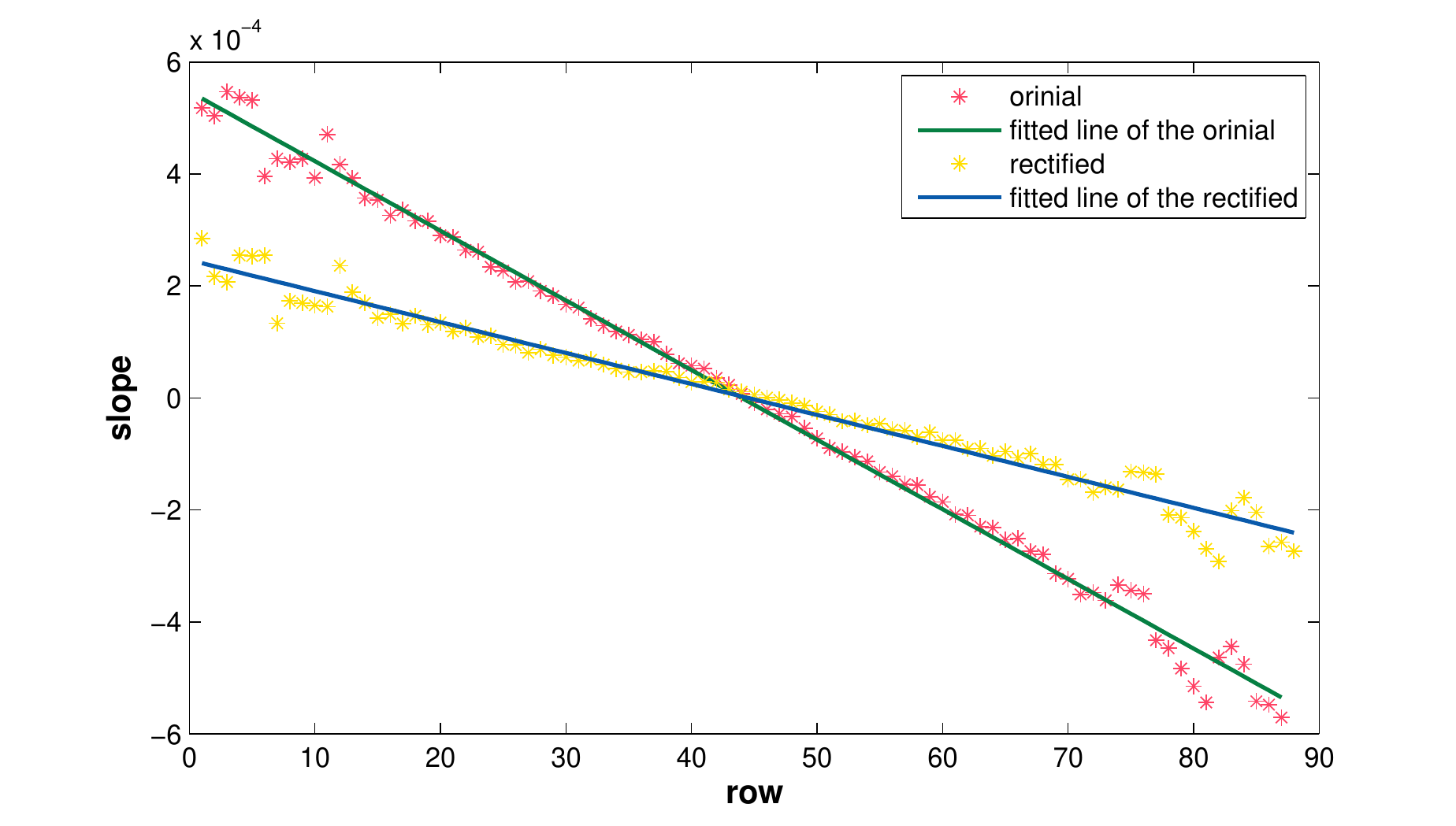}
\caption{
The slopes fitted by 88 line of micro-lens images' centers from the original and rectified white image of a physical focused plenoptic camera. Every slope is fitted by 115 centers from the same row of micro-images.}
\label{fig:slopes_centers}
\end{figure}

The centers of the micro images $\left(x_g^\prime, y_g^\prime \right)\!\!^\mathrm{T} $ are recognized by a raw image shooting a white scene or other pure color \cite{cho2013modeling,Chunping2016Decoding}. The misalignment of the MLA makes the diameter of the micro-lens image non-uniform. As shown in Fig.\ref{fig:slopes_centers}, the slopes fitted by a line of the centers of micro-lens images are descending linearly. The rate of descending is related to the rotation of the MLA. The range of the slopes of the rectified images is smaller than the slopes of the original images. It indicates that the diameters of the rectified micro images are more uniform. Taking the misalignment of the MLA into account \cite{thomason2014calibration}, we rectify the raw image by a homography to make the micro-lens images uniform. More importantly, by Eq.\ref{eq_RtXg} and Eq.\ref{eq_xg_}, a homography with 8 degree of freedom is sufficient to preprocess the raw image, thus the two non-parallel light field planes are reparameterized to parallel planes. The experiments in Section~\ref{subsec:simu} and \ref{subsec:physical} are based on the rectification.

\subsection{Simulated data}
\label{subsec:simu}

We first verify the calibration method on the simulated images rendered in MATLAB. The image sensor resolution is $4008\times2672$ with 9 $\mu m$ pixel width. The focused plenoptic camera consists of a main lens with 50 $mm$ focal length and a MLA with 300 $\mu m$ diameter and 2.726 $mm$ focal length in hexagon layout. The calibration board is a pattern with $5\times5$ points of $54.0\times54.0 \, mm$ cells. We render 12 raw images with different poses of the board.

For a focused plenoptic camera, to recognize the multiple projections of the same scene point, we preprocess the raw image using a white image and then use template matching by normalized cross-correlation (NCC).

\begin{figure}
\centering
\includegraphics[width=105mm]{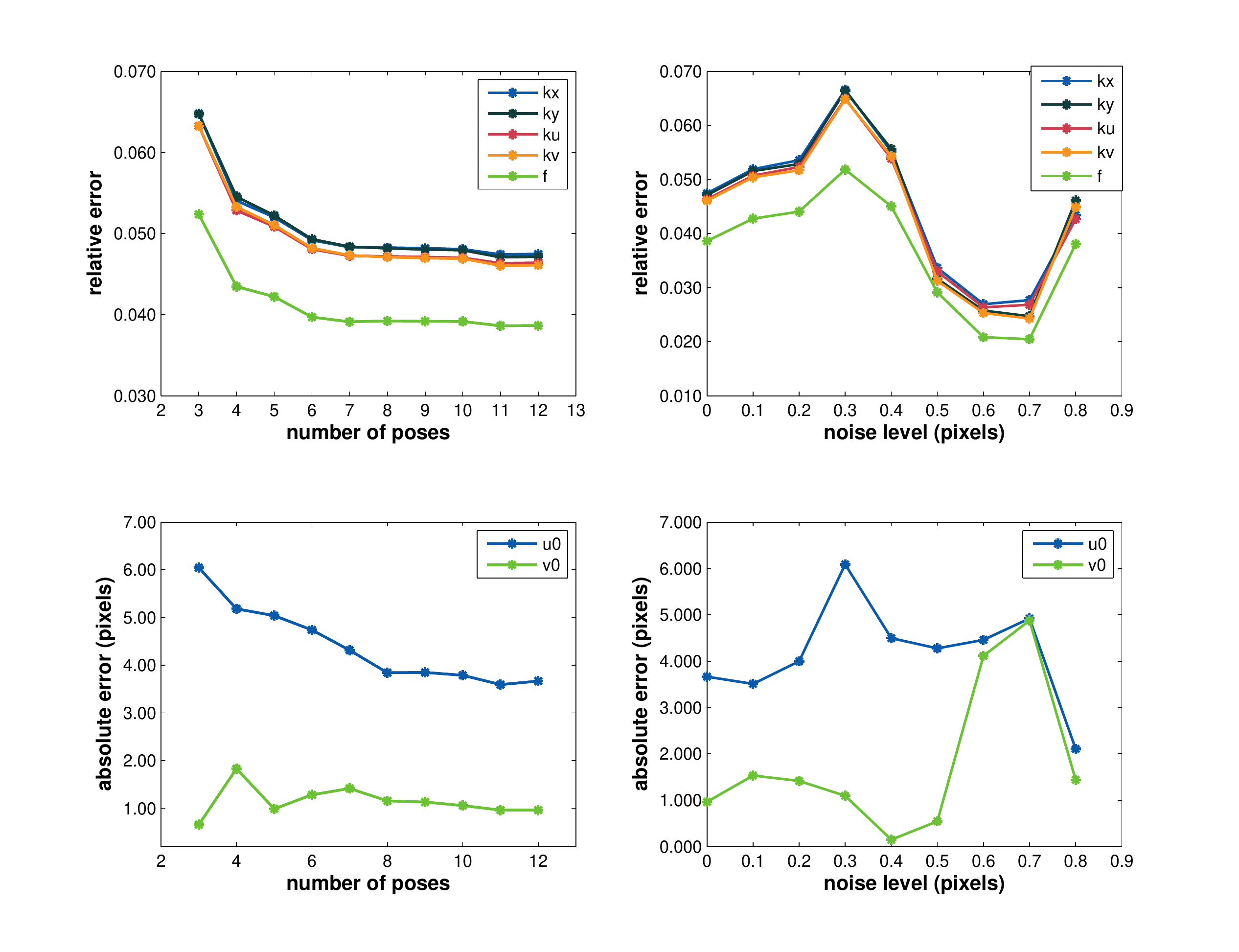}
\caption{
The calibration errors on simulated data with different numbers of poses (the left-two) and different noise levels (the right-two).}
\label{fig:SimuResults}
\end{figure}

We test the performance with respect to the numbers of poses of the calibration board. We vary the number of images from 3 to 12. The calibration results with increasing number of poses are shown in Fig.\ref{fig:SimuResults}. The errors decrease when more images are used. From to 3 to 4, the errors of most parameters decrease significantly. When the number of poses is more than 6, all parameters tend to be stable. In addition, the ground truth are calculated by the input paramters of the simulation and the equations in Section~\ref{subsec:tpp2}. 

In addition, we add the projections of the total 12 raw images with Gaussian noise with 0 mean and standard deviation varied from 0.1 pixels to 0.8 pixels. The results are shown in Fig.\ref{fig:SimuResults}. Due to that there are at least 12 projections of the same scene point in single raw image, the calibration results are still reasonable with different noise levels. It verifies the robustness of the calibration method.

\subsection{Physical camera}
\label{subsec:physical}

\vspace{-3mm}
\begin{figure}
\centering
\includegraphics[width=90mm]{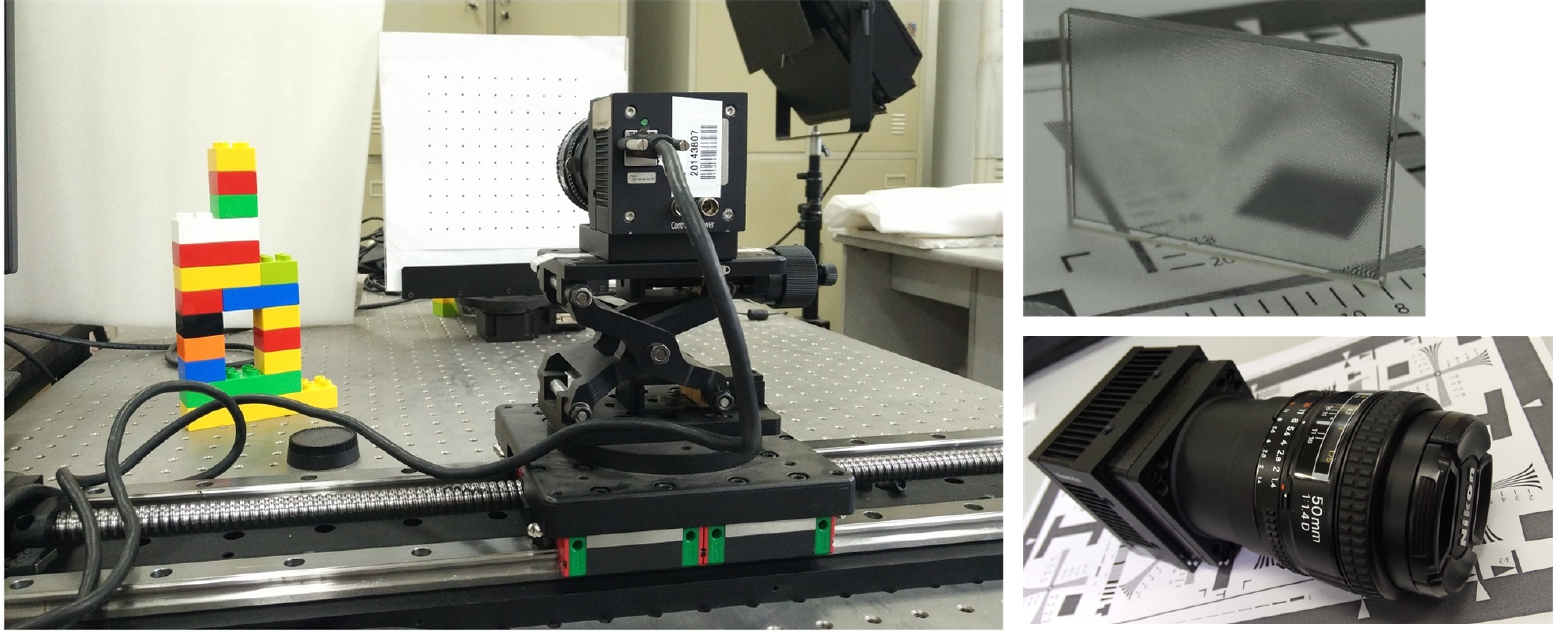}
\vspace{-2mm}
\caption{
The self-assembly focused plenoptic camera and the MLA inside the camera.}
\label{fig:physicalCamera}
\end{figure}

We capture raw images of calibration board and real scene using a self-assembly focused plenoptic camera. The camera and the MLA are shown in Fig.\ref{fig:physicalCamera}. The camera consists of a GigE camera with a CCD image sensor whose resolution is $4008\times2672$ with 9 $mm$ pixel width, a Nikon AF Nikkor f/1.4D F-mount lens with 50 $mm$ focal length, and a MLA with 300 $\mu m$ diameter and 2.726 $\mu m$ focal length in hexagon layout. The calibration board is $10\times10$ points with 2$cm$ $\times$ 2 $cm$ cells (Fig.\ref{fig:physicalCamera}). We shoot 9 raw images with different poses of the board, and 9 raw images with the same real scene and a calibration board. All the raw images are preprocessed by the method mentioned in Section~\ref{subsec:rect}.

\begin{table}[tbp]
\caption{Calibration results of a physical camera with different numbers of poses.}
\centering
\tiny
\label{cp_tab4}
\begin{tabular}{c|c|c|c|c|c|c|c|c}
\hline
Parame & \multicolumn{2}{|c|}{3 poses} & \multicolumn{2}{|c|}{5 poses} & \multicolumn{2}{|c|}{7 poses} & \multicolumn{2}{|c}{9 poses} \\
\cline{2-9}
-ter & $4\times 5$  & $8\times10$ & $4\!\times\! 5$ & $8\!\times\! 10$ & $4\!\times\! 5$  & $8\!\times\! 10$ & $4\!\times\! 5$  & $8\!\times\! 10$ \\
\hline
$k_x$  & 31.1525 & 32.2787 & 29.0403 & 28.7948 & 27.5295 & 27.7046 & 27.1743 & 27.0686\\
$k_y$  & 30.9687 & 32.1397 & 29.0474 & 28.8172 & 27.5140 & 27.6846 & 27.1752 & 27.0632\\
$k_u$  & 1230.23 & 1271.44 & 1152.98 & 1143.66 & 1097.09 & 1103.30 & 1084.01 & 1079.72\\
$k_v$  & 1224.51 & 1267.02 & 1153.47 & 1144.48 & 1096.8 & 1102.77 & 1084.28 & 1079.66\\
$u_0$/pixel  & -3364.99 & -3598.28 & -3440.39 & -3604.41 & -3475.55 & -3685.79 & -3508.18 & -3691.41\\
$v_0$/pixel  & -7162.50 & -7417.96 & -6960.49 & -7075.08 & -6891.50 & -7043.57 & -6867.73 & -6987.77\\
$f$/pixel    & 31098.7 & 31605.3 & 30309.7 & 30162.6 & 29878.2 & 30012.6 & 29742.5 & 29701.1\\
\hline
$s_1$  & 0 & 0 & 0 & 0 & 0 & 0 & 0 & 0\\
$s_2$  & 0 & 0 & 0 & 0 & 0 & 0 & 0 & 0\\
$t_1$  & -2.44e-13 & -2.54e-13 & -3.81e-13 & -3.84e-13 & -3.85e-13 & -3.85e-13 & -3.72e-13 & -3.80e-13\\
$t_2$  & 1.79e-22 & 1.60e-22 & 3.059e-22 & 2.89e-22 & 3.65e-22 & 3.32e-22 & 3.69e-22 & 3.47e-22\\
\hline
RMS    & 0.30394 & 0.31505 &  0.27223 & 0.29952 & 0.26344 & 0.29151 & 0.24877 & 0.27846\\
\hline
\end{tabular}
\label{tab:RealResult}
\vspace{-2mm}
\end{table}

\begin{figure}[tbp]

\centering
\subfigure[\,]{
\label{fig:RealResultPoses}
\includegraphics[width=0.55\textwidth]{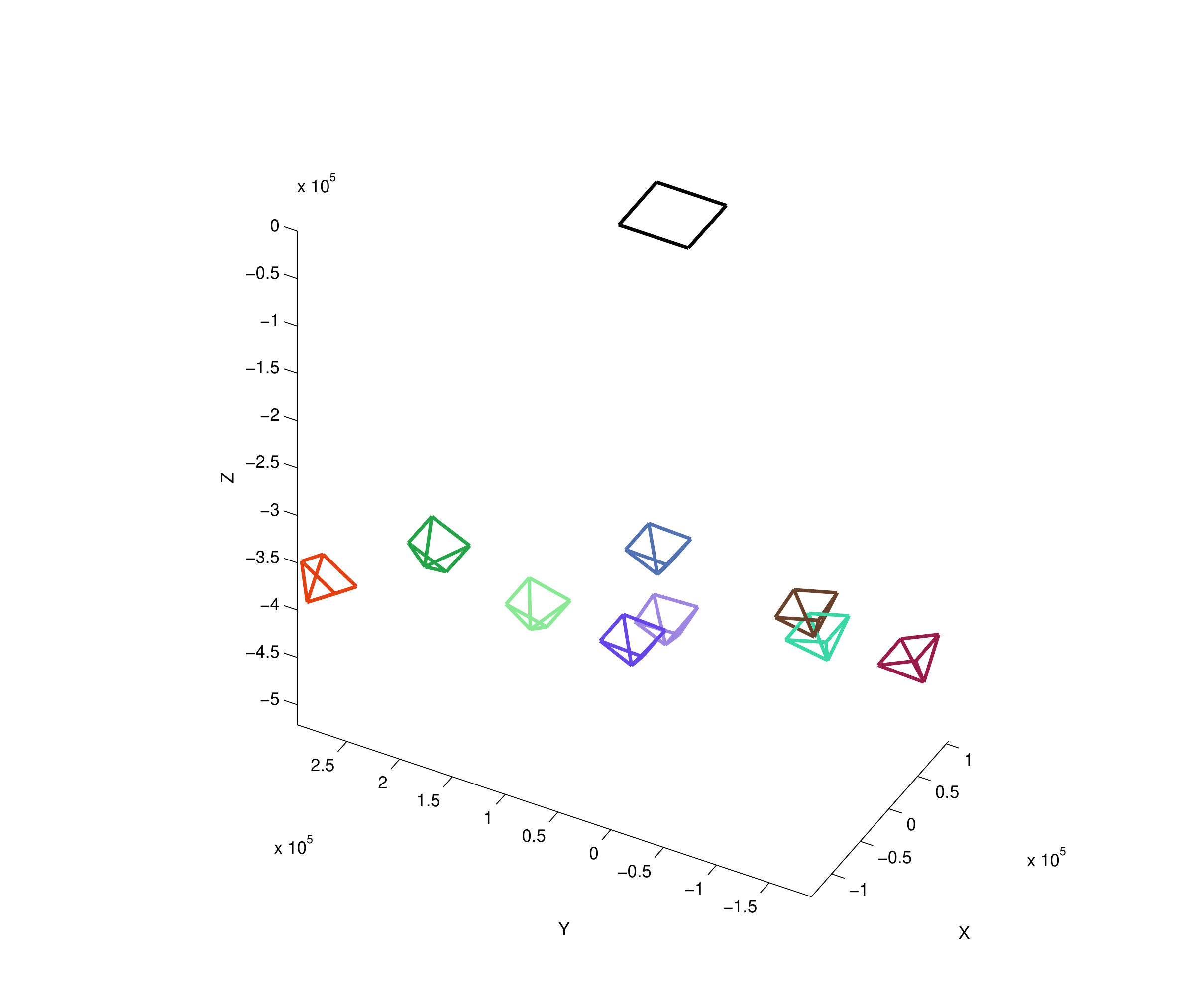}}
\vspace{-1mm}
\subfigure[\,]{
\label{fig:RealResultErrors}
\includegraphics[width=0.42\textwidth]{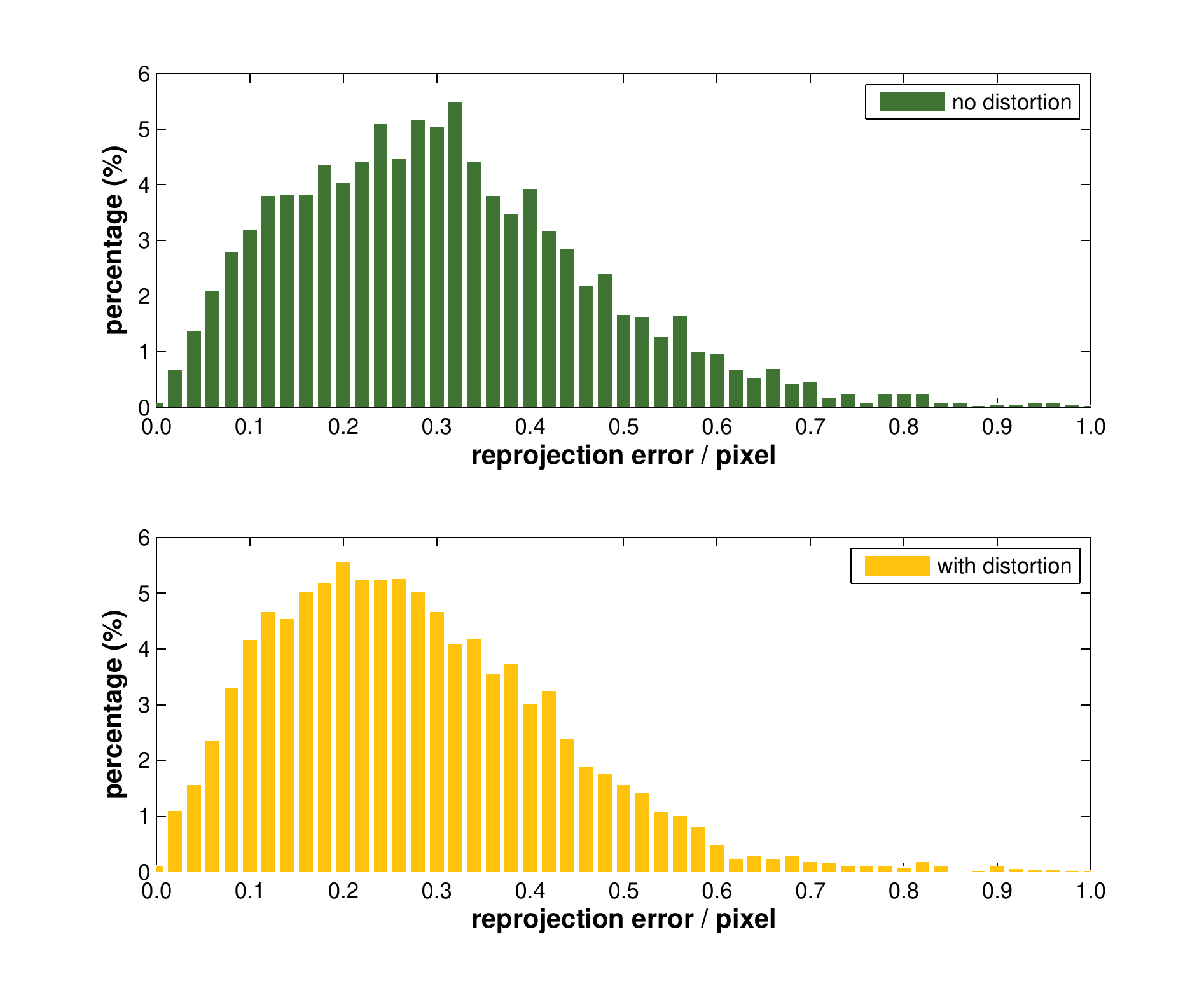}}
\caption{
(a) shows the estimated poses of the 9 raw images with a calibration board (the black parallelogram on the top). (b) shows the histograms of the distribution of the re-projection error. The histograms are calculated without or with the distortion coefficients, and the mean errors are 0.31623 and 0.27846 pixels respectively.}
\label{fig:RealResult}
\end{figure}

The results of the estimated intrinsic parameters are listed in Tab.1. We apply our calibration method to the first 3, 5, 7, and all 9 raw images in different poses. As shown in Tab.\ref{tab:RealResult}, the re-projections RMS error is less than 0.3 pixels when at least 5 poses are used. The RMS errors are quite consistent with different numbers of poses. With the same number of poses, the parameters are close when the number of the calibration points is changed from $4\times 5$ points to $8\times 10$ points. The poses estimated using the 9 raw images with a calibration board and the histogram of the re-projection errors are shown in Fig.\ref{fig:RealResult}. Fig.\ref{fig:RealResultErrors} shows that most of the errors are less than 0.5 pixels, and the re-projection errors decrease with the optimization on distortion.

\begin{figure}
\vspace{-2mm}
\centering
\includegraphics[width=120mm]{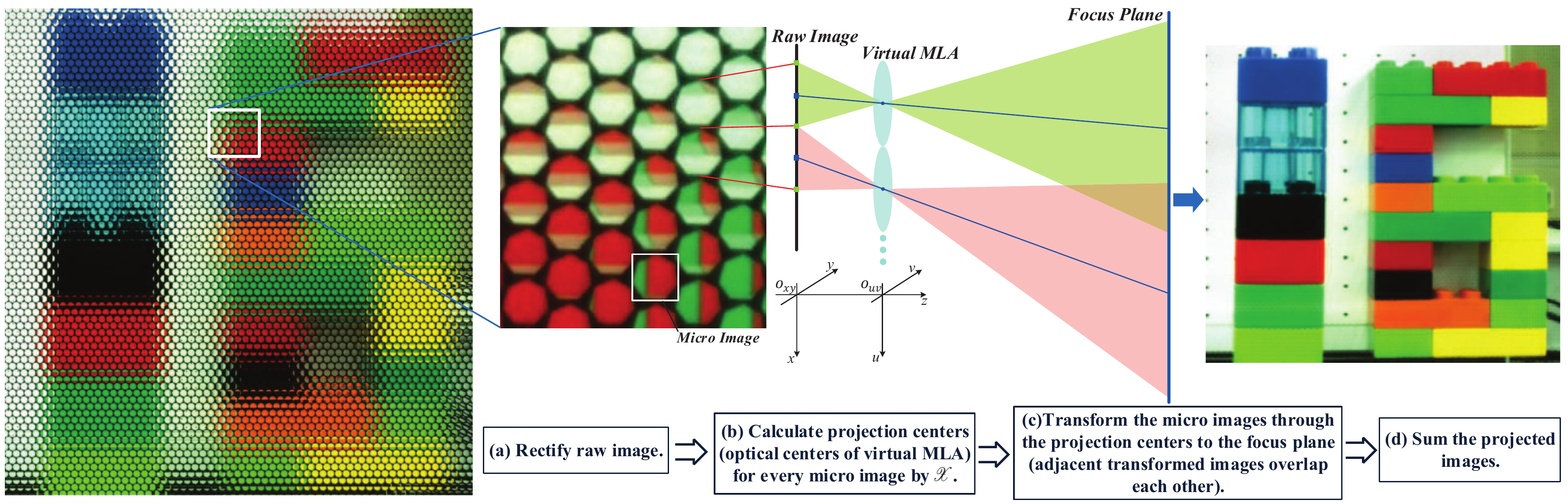}
\caption{
The refocus rendering pipeline of a focused plenoptic camera.}
\label{fig:RenderPipeline}
\end{figure}

\begin{figure}
\centering
\subfigure[\,]{
\label{fig:RenderImgsSub1}
\includegraphics[width=0.52\textwidth]{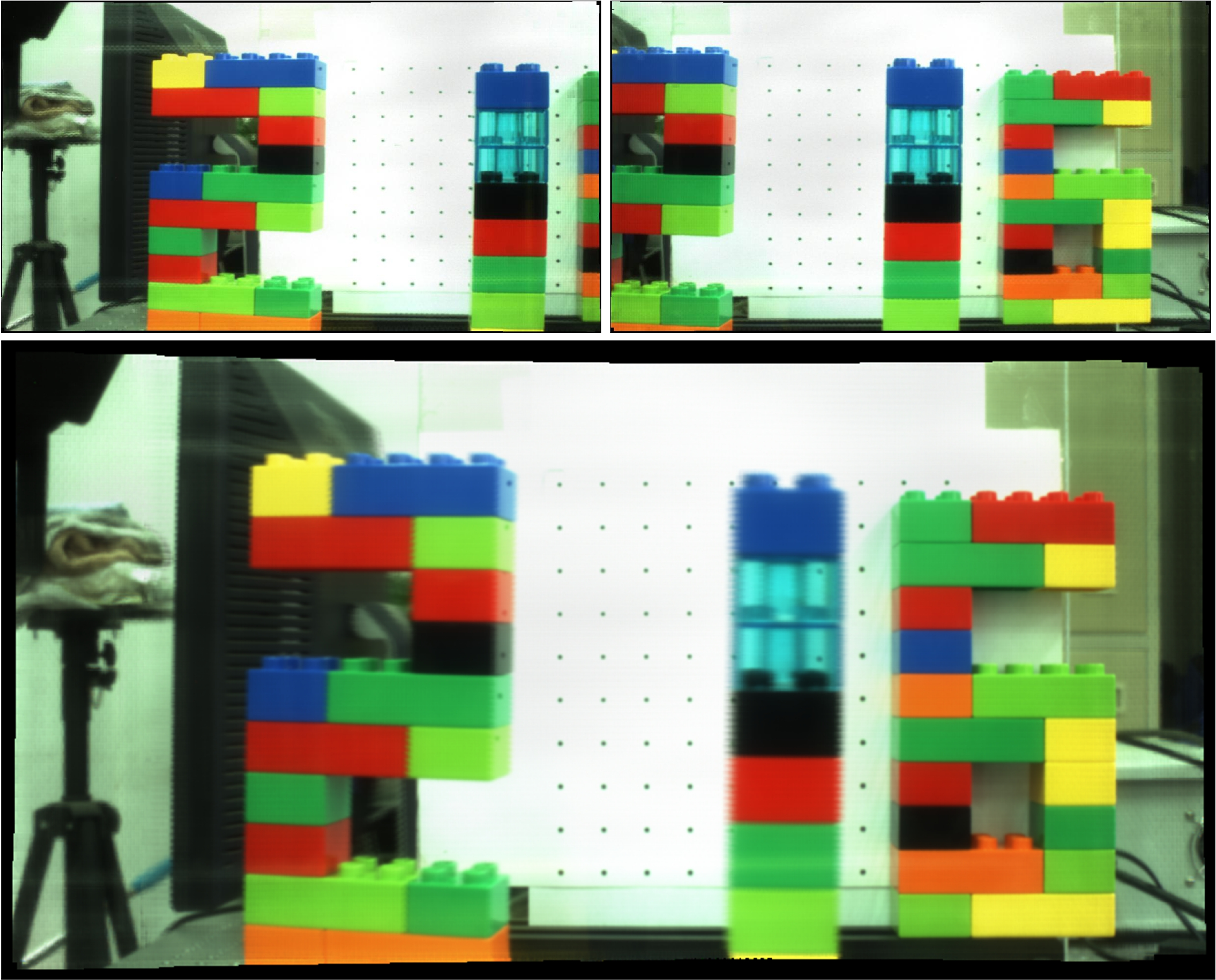}}
\subfigure[\,]{
\label{fig:RenderImgsSub2}
\includegraphics[width=0.45\textwidth]{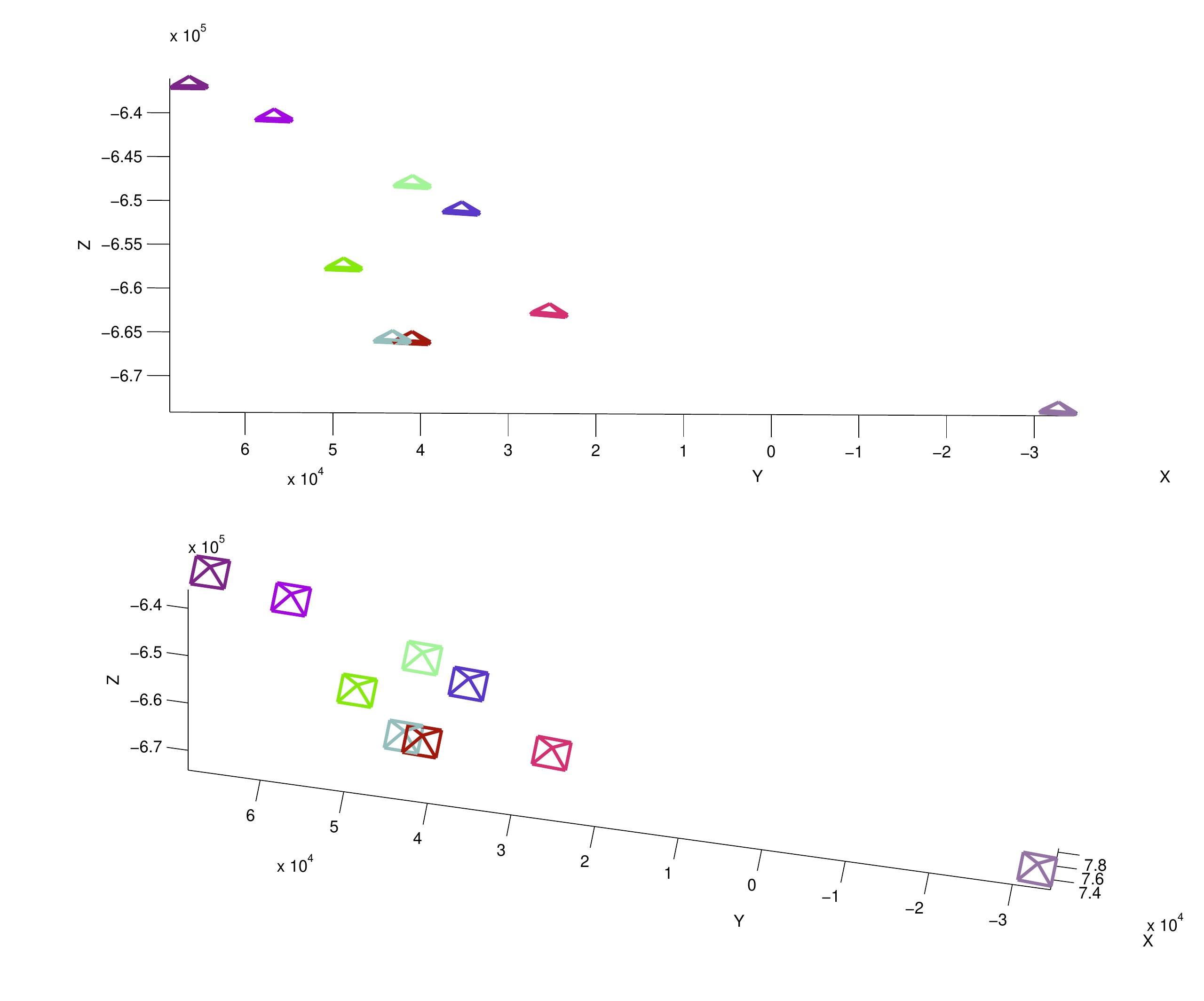}}
\vspace{-4mm}
\caption{
(a) shows the rendered images from the physical focused plenoptic camera. The top are the most left view and the most right view and the bottom is the stitched image of the different 9 poses. (b) shows two of the views of the estimated posed from the 9 raw images.}
\label{fig:RenderImgs}
\end{figure}

After the estimation of $\mathscr{X}$, the virtual optical centers of the MLA are calculated, thus the rays' directions are obtained \cite{perwass2012single}. The refocus rendering pipeline of a focused plenoptic camera is shown in Fig.\ref{fig:RenderPipeline}. The rendered images by ray tracing using 9 raw images in different poses are shown in Fig.\ref{fig:RenderImgs}.

\section{Conclusion}
\label{sec:Conclusion}

In this paper, we present a novel unconstrained TPP model to describe the relationship between the 4D rays and the 3D scene structure by a projective transformation. To calibrate the focused plenoptic camera, we simplify the imaging system as a 7-parameter TPP model. Compared with the previous calibration method on the focused plenoptic cameras, we substitute the refraction of the main lens and simplify the image system as a 7-intrinsic-parameter unconstrained TPP model, which is closer to the optical path and the imaging principles. We derive the closed-form solution for the intrinsic and extrinsic parameters and then refine the parameters by minimizing the re-projections error via LM algorithm. Both simulated data and real data verify the robustness and validity of our proposed method. Due to the image features, the multiple projections are more convenient to be recognized in a focused plenoptic camera. The recognition of projections in a conventional plenoptic camera is mentioned in the work \cite{bok2014geometric,bergamasco2015adopting}. Moreover, the TPP coordinate system for the conventional plenoptic camera can be regarded as the main lens plane and the focal plane of MLA outside the camera. Therefore, the calibration for a conventional plenoptic camera can be completed by estimating the view coordinates, the focal length and the principle points of the sub-aperture images.

\bibliographystyle{plain}
\bibliography{egbib}

\end{document}